\documentclass[11pt]{article}

\usepackage[final]{acl}

\usepackage{times}
\usepackage{latexsym}

\usepackage{pifont}
\usepackage[T1]{fontenc}

\usepackage[utf8]{inputenc}

\usepackage{microtype}

\usepackage{inconsolata}
\usepackage{multirow}
\usepackage{amsmath}
\usepackage{amsfonts}

\usepackage{graphicx}
\usepackage{algorithm}
\usepackage{algpseudocode}

\usepackage{subcaption}
\usepackage{xspace}
\newcommand{\ours}{OWL\xspace}
\newcommand{\hybridours}{HOWL\xspace}
\newcommand{\ourbench}{LongSpecBench\xspace}
\newcommand{\ourtoken}{[SPEC]\xspace}

\usepackage{etoolbox} %

\newtoggle{showprev}

\newtoggle{shownew}
\toggletrue{shownew}

\usepackage{array} %
\makeatletter
\newcommand*{\@rowstyle}{}
\newcommand*{\rowstyle}[1]{%
  \gdef\@rowstyle{#1}%
  \@rowstyle\ignorespaces%
}
\newcolumntype{=}{%
  >{\gdef\@rowstyle{}}%
}
\newcolumntype{+}{%
  >{\@rowstyle}%
}
\makeatother

\definecolor{Emerald}{RGB}{192, 41, 66}
\iftoggle{shownew}{
    
    \newcommand{\jsleeobj}[1]{ %
    {\let\Cap\caption
    \def\caption##1{\Cap{\color{Emerald}##1}}
    \color{Emerald}#1}
    }
}
{
    
    \newcommand{\jsleeobj}[1]{#1}
    
}

\definecolor{Gray}{RGB}{200, 200, 200}
\iftoggle{showprev}{
    \newcommand{\jsleeprev}[1]{{\color{Gray}#1}}
    \newcommand{\jsleeprevobj}[1]{ %
    {\let\Cap\caption
    \def\caption##1{\Cap{\color{Gray}##1}}
    \color{Gray}#1}
    }
    \newcommand{\jsleeprevrow}[1]{\rowstyle{\color{Gray}}#1}
}
{
    \newcommand{\jsleeprev}[1]{}
    \newcommand{\jsleeprevobj}[1]{}
    \newcommand{\jsleeprevrow}[1]{}
}

\title{
 \includegraphics[height=2em]{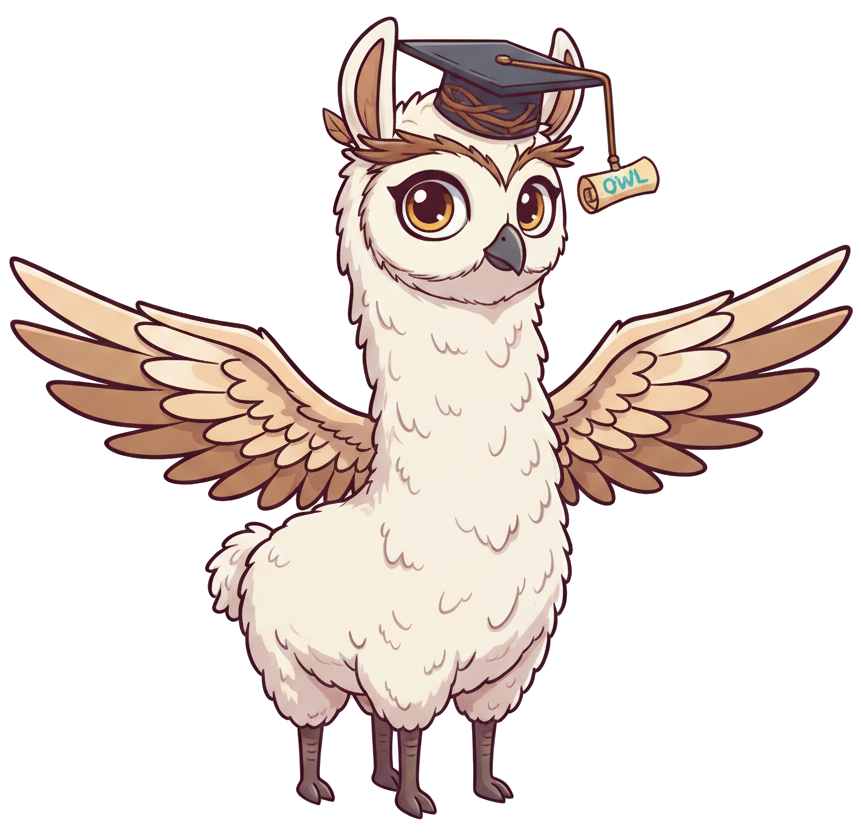} 
\ours: Overcoming Window Length-Dependence in\\Speculative Decoding for Long-Context Inputs}

 \author{\textbf{
Jaeseong Lee\textsuperscript{*},
Seung-won Hwang\thanks{Work done while visiting Snowflake. Correspond to seungwonh@snu.ac.kr},
Aurick Qiao,}\\
\textbf{Gabriele Oliaro\textsuperscript{\textdagger},
Ye Wang,
Samyam Rajbhandari}
\\
Snowflake AI Research, Seoul National University\textsuperscript{\rm *}, Carnegie Mellon University\textsuperscript{\rm \textdagger}\\
 }

\begin{document}
\maketitle
\begin{abstract}
Speculative decoding promises faster inference for large language models (LLMs), yet existing methods fail to generalize to real-world settings. Benchmarks typically assume short contexts (e.g., 2K tokens), whereas practical workloads involve long contexts. We find current approaches degrade severely with long contexts; for instance, EAGLE3 even slows down the generation speed by 0.81\texttimes.
We address these limitations by releasing a new long-context benchmark (\ourbench) and introducing a novel model (\ours). \ours achieves about 5\texttimes\xspace higher acceptance length than EAGLE3 on long-context inputs through three innovations: (1) an LSTM-based drafter conditioned only on the last-token state, making it generalize to various lengths, (2) a special token \ourtoken in the verifier that produces richer representation for drafter, and (3) a hybrid algorithm combining both tree and non-tree decoding methods. We release all code and datasets to advance future research.\footnote{\url{https://anonymous.4open.science/r/owl-BFB8}}
\end{abstract}
\section{Introduction}
Long-context generation is increasingly being important-- Use cases include reasoning with long thinking path~\citep{DeepSeekR12025deepseek-ai}, multi-turn conversations, and agentic systems~\citep{MagicDec2025sadhukhan}.
In response, large language models (LLMs) have rapidly evolved to handle much longer input contexts-- originally from 2K tokens by Vicuna~\citep{vicuna2023} to 128K by Llama-3~\citep{Llama3Llama2024dubey}.
This expanded context capability unlocks complex reasoning and comprehensive information access, but it also comes with a steep computational cost: generating each token becomes slower as context length grows, due to the sequential nature of autoregressive decoding and the large memory that must be accessed at every step. 

Speculative decoding~\citep{specdecFast2023leviathan,Sequoia2024chen,SpecInfer2024miao,Medusa2024cai,EAGLE2024li,EAGLE22024li,EAGLE32025li,SuffixDecoding2025oliaro,SAM2025hu,Turning2025luo} is a promising solution to accelerate LLM inference. It uses a faster drafter to predict several upcoming tokens in a tree or sequence structure, and let the target LLM verify the drafted tokens. The tree- or non-tree decoding algorithm will accept proper tokens to keep the output distribution, which is used by the drafter to generate another set of next tokens. In memory-bound scenarios, the cost of verifying the multiple drafted tokens is hidden, leading to significant speedups-- State-of-the-art speculative decoding method, EAGLE3~\citep{EAGLE32025li} claims 6.5\texttimes \xspace speedup over standard decoding.

However, such speedups often fail to generalize to real-world scenarios. First, common speculative benchmarks mostly assume a short 128 context, 2K context at maximum (e.g., Vicuna), and a batch size of one. 
However, in a real-world workload, such a short context easily shifts out of the memory-bound regime by increasing the batch size~\citep{Synergy2023su,SpecInfer2024miao,TurboSpec2025liu,EAGLE32025li}. Benchmarks with much longer contexts are more realistic for speculative decoding.

\begin{figure*}[]
\centering
\begin{minipage}[t]{.48\linewidth}
    \includegraphics[width=\textwidth]{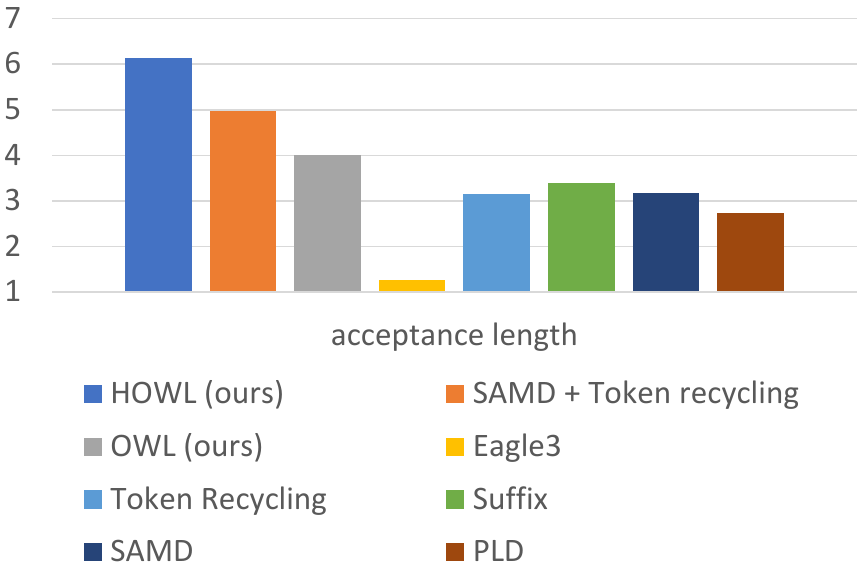}
    \subcaption{Llama3.1-8B-Instruct}\label{fig:llama31_accept}
\end{minipage}  
\begin{minipage}[t]{.48\linewidth}
    \includegraphics[width=\textwidth]{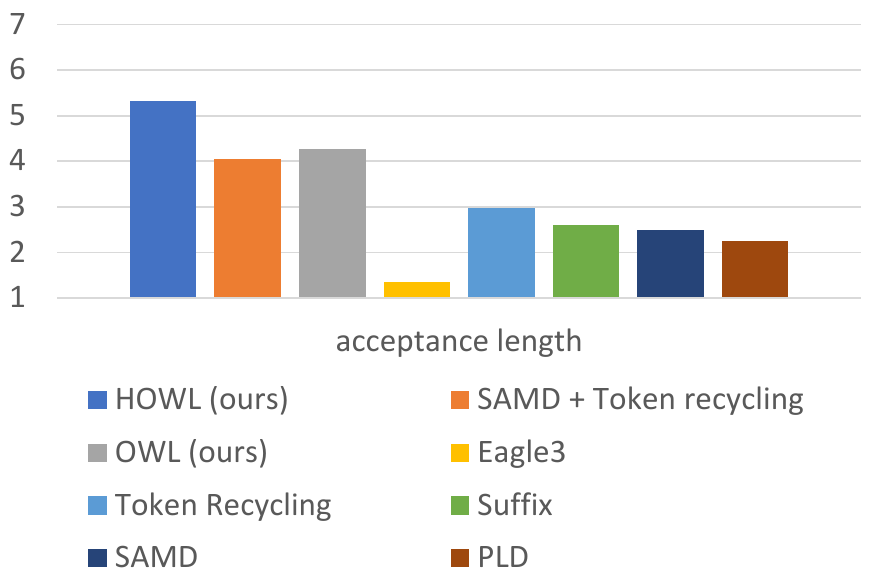}
    \subcaption{Llama-3.3-70B-Instruct}\label{fig:llama33_accept}
\end{minipage}  
\caption{Speculative decoding performance on long-context inputs on Llama-3.1-8B-Instruct model on \ourbench. The acceptance length is the number of accepted tokens per verification step with the target LLM. Higher is better. While EAGLE3 (yellow) fails to generalize to long-context inputs, our method (blue) can achieve almost 5\texttimes \xspace higher acceptance length than EAGLE3.}
\label{fig:accept}
\end{figure*}

Second, existing methods degrade sharply in long contexts. For example, EAGLE3 achieves an acceptance length of only 1.28 (\autoref{fig:llama31_accept}), generating just 0.28 extra tokens per verification step.

\begin{figure}
\centering
\includegraphics[width=\columnwidth]{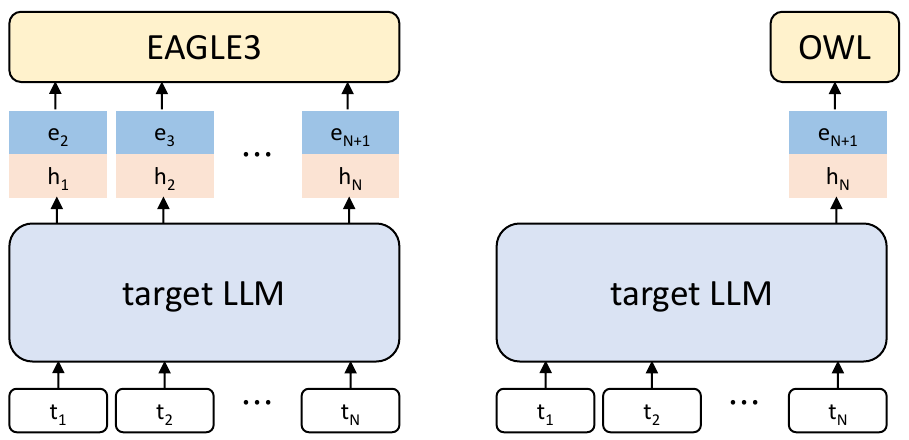}
\caption{EAGLE3 (left) and \ours (right).}
\label{fig:eagle3_vs_ours}
\end{figure}

To address these issues, we introduce both a new long-context benchmark (\ourbench) and a new model (\ours). \ours achieves almost 5\texttimes \xspace higher acceptance length than EAGLE3 on long-context inputs, with innovations in each key component of speculative decoding-- drafter, verifier, and decoding algorithm:
\begin{itemize}
\item Length-general drafter: Unlike EAGLE3's transformer, which fails to generalize beyond its trained 2K window~\citep{Long2021tay}, \ours uses an LSTM drafter conditioned only on the last-token state, avoiding context-length dependence (\autoref{fig:eagle3_vs_ours}).

\item Specialized token for verifier: We introduce \ourtoken \xspace to provide a richer representation to the drafter, by letting the verifier predict an additional token beyond the verified tokens.

\item Hybrid algorithm of tree- and non-tree decoding: We combine our tree-decoding-based model with an existing non-tree decoding method, achieving higher acceptance length and speedups.

\end{itemize}

First, we attribute the failure of existing EAGLE3 models to the transformer architecture~\citep{transformersAttention2017vaswani} they use. The transformers do not generalize well beyond their seen context window, which EAGLE3 set
as 2K following the common benchmarks. One possible approach would training a transformer with long-context. 
However, it requires a special dataset with long-context. Additionally, we report EAGLE3-L, even after such training, is inferior to our proposal (\autoref{tab:longeagle}).

Instead, \ours removes the need to feed all input tokens to the drafter. As depicted in \autoref{fig:eagle3_vs_ours}, we rely on the hidden state of a single token, the last token only, to generate the next tokens. To materialize it, we leverage the LSTM~\citep{lstmLong1997hochreiter} architecture. This design makes the drafter agnostic to the context length, and thus length-generalized.

Second, we introduce a special token in the verifier, to provide the drafter richer representation.
We let the target LLM try to estimate an additional token beyond the accepted tokens at the verification step. We materialize this idea by appending a special token \ourtoken to the input (\autoref{fig:inference_overview}). With the hidden state generated from \ourtoken, we can increase the acceptance length of our drafter further, even keeping the latency intact.

\begin{figure*}[]
\centering
\begin{minipage}[t]{.35\textwidth}
    \includegraphics[width=\textwidth]{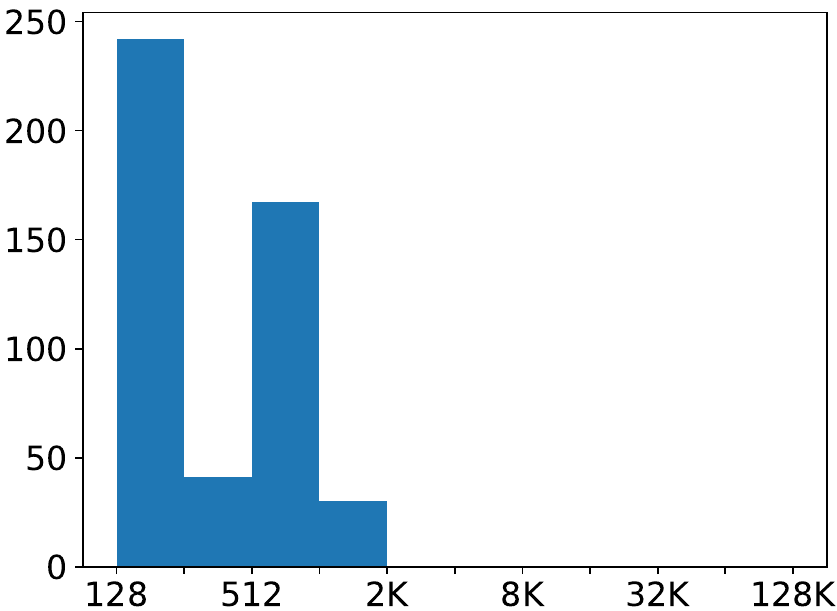}
    \subcaption{SpecBench~\citep{SpecBenchUnlocking2024xia}}\label{fig:specbench_len}
\end{minipage}  
\begin{minipage}[t]{.35\textwidth}
    \includegraphics[width=\textwidth]{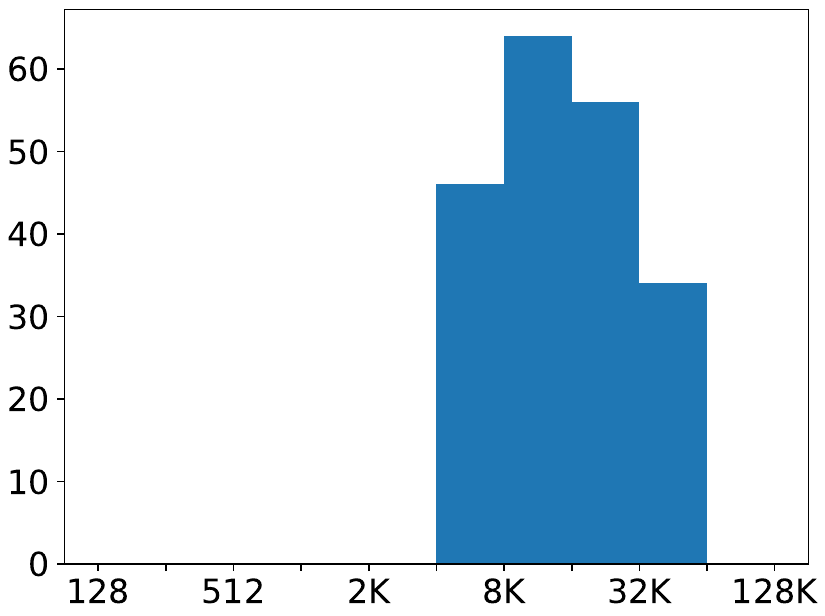}
    \subcaption{\ourbench (ours)}\label{fig:ourbench_len}
\end{minipage}  
\caption{
Context length distribution of existing benchmarks and \ourbench. Existing benchmarks, such as SpecBench~\citep{SpecBenchUnlocking2024xia}, focus on short-context inputs (within 2K tokens). In contrast, \ourbench (ours) contains long-context inputs, with a length of tokens up to 64K tokens.}
\label{fig:len}
\end{figure*}

Lastly, we explore the complementary benefit of tree- and non-tree decoding-based speculative decoding methods. We find that while some non-tree decoding based methods, such as SuffixDecoding~\cite{SuffixDecoding2025oliaro}, have lower acceptance length than our tree-decoding based methods, they can provide extremely high acceptance length in some cases (\autoref{fig:hist}). Since the best case scenario of non-tree decoding is accepting all the sequence of drafted tokens, we conditionally leverage this to enhance our best case scenario.

Experiments on variant scales of LLMs, such as Llama-3.1-8B-Instruct and Llama3.3-70B-Instruct~\citep{Llama3Llama2024dubey}, demonstrate the effectiveness of our method.
We publicly release our code and datasets to facilitate future research.\footnotemark[\value{footnote}]

\section{Related Work}
Speculative decoding~\cite{specdecFast2023leviathan} is a lossless acceleration technique of LLM decoding.
It speculates multiple tokens in a row with a fast drafter, and verifies the tokens with the target LLM in parallel. In a memory-bound scenario, such additional computational cost is hidden.

While the initial technique speculated tokens in a sequence-like structure, tree-decoding emerged to speculate tokens in a tree-like structure~\cite{SpecInfer2024miao,Sequoia2024chen,Medusa2024cai,EAGLE2024li}, although some recent works~\cite{SuffixDecoding2025oliaro} proposed non-tree decoding methods that perform comparably.

Recently, EAGLE3~\cite{EAGLE32025li} achieved the state-of-the-art in benchmarks such as SpecBench~\cite{SpecBenchUnlocking2024xia}. In response, inference engines such as vLLM~\citep{vllmEfficient2023kwon} or SGLang~\citep{SGLang2024zheng} also integrated EAGLE3 as a built-in speculative decoding to accelerate LLM inference.

However, we find that the high acceptance length of EAGLE3 is not viable when the context length is beyond the trained context window. We design a benchmark to consider such a scenario, and propose a length-generalized speculative decoding method.

\section{Proposed Method}
\subsection{\ourbench: A New Benchmark with Long-Context Inputs}
Existing benchmarks for speculative decoding, such as SpecBench~\citep{SpecBenchUnlocking2024xia} or the EAGLE3 benchmark~\citep{EAGLE32025li}, primarily focus on short-context inputs, typically within 2K tokens (\autoref{fig:specbench_len}, \ref{fig:eagle3_len}). However, real-world applications often involve much longer contexts, which can significantly impact the performance of speculative decoding methods. 

To address this gap, we build \ourbench, a new benchmark specifically designed to evaluate the effectiveness of speculative decoding techniques on long-context inputs.

To utilize the real-world use cases, we leverage WildChat-4.8M,\footnote{hf.co/datasets/allenai/WildChat-4.8M} which records conversations between human users and ChatGPT~\citep{WildChat2024zhao}. We sample 200 examples with input length distributed ranging from 4K to 64K tokens (\autoref{fig:ourbench_len}).

Surprisingly, this new benchmark reveals that existing speculative decoding methods, such as EAGLE3, do not generalize well to long-context inputs. For instance, EAGLE3 fails to produce a high acceptance length on \ourbench-- only 1.28 (\autoref{fig:llama31_accept}). This highlights the need for new approaches that can effectively handle long-context scenarios.

\begin{figure*}[]
\centering
\includegraphics[width=0.9\linewidth]{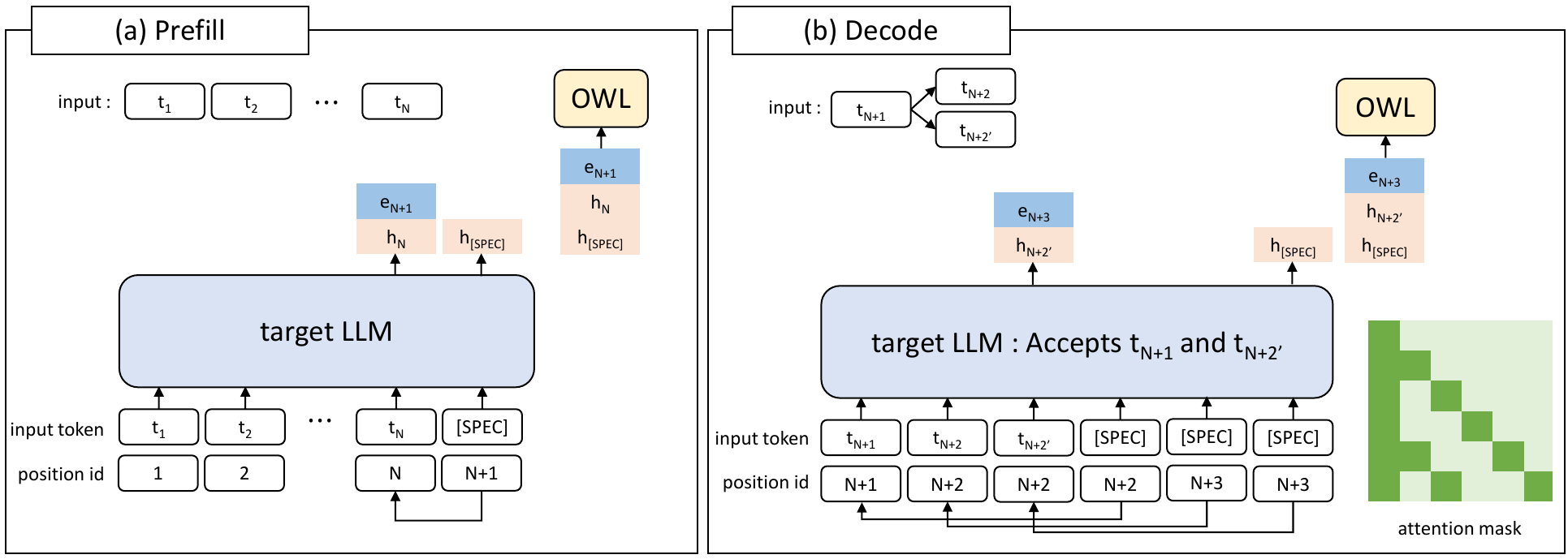}
\caption{
Overview of \ourtoken for verifier in the inference stage, each for (a) prefill stage and (b) decode stage.}
\label{fig:inference_overview}
\end{figure*}

\begin{figure}[]
\centering
\includegraphics[width=\linewidth]{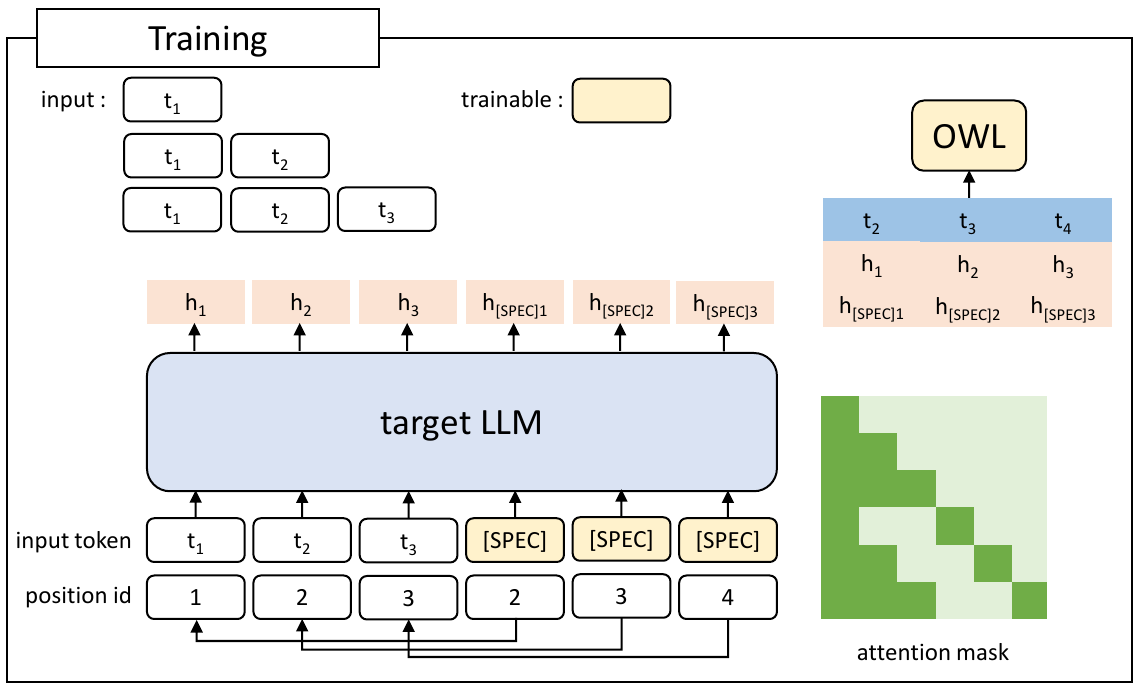}
\caption{
Overview of \ourtoken for verifier in the training stage.}
\label{fig:training_overview}
\end{figure}

\subsection{Length-Generalized Speculative Decoding When Input Exceeds Trained Length}
In this section, we describe our proposed method, consisting of innovation in each of three key components of speculative decoding: drafter, verifier, and decoding algorithm. 
First, we propose a length-generalized drafter to address the context-length dependence of existing transformer-based drafters (\S \ref{sec:length_invariant}). 
Second, we introduce a specialized token for the verifier to signal the target LLM to predict beyond the verified tokens (\S \ref{sec:spec_token}). 
Lastly, we present a hybrid algorithm that combines tree-decoding and non-tree decoding methods to leverage the strengths of both (\S \ref{sec:hybrid}).

\subsubsection{\ours: Length-Generalized Drafter}\label{sec:length_invariant}

As transformer architecture is dependent on the trained context window~\cite{Long2021tay,Qwen25_1MQwen22025yang},
EAGLE3, following the transformer architecture, has similar limitations.

In contrast, we propose \ours, a length-generalized drafter by removing the dependency on all the input tokens. 
As depicted in \autoref{fig:eagle3_vs_ours}, unlike EAGLE3, which feeds all input tokens to the drafter, \ours relies only on the hidden state of a single token, the last token only, to speculate the next tokens. 
To materialize it, we leverage the LSTM~\citep{lstmLong1997hochreiter} architecture. This design makes the drafter agnostic to the context length, and thus length-generalized.
The ability to understand the long context would be yielded to the target LLM~\citep{Llama3Llama2024dubey}.
Empirically, we find that even a short context length, such as 256 tokens, is sufficient to train such a drafter to support long-context inputs.

In detail, we establish the architecture of \ours as follows. Given the predicted next token $t_{N+1}$, the last hidden states $h_{N} \in \mathbb{R} ^{d_0}$, and a trainable embedding layer $E$, we first sum up embedding of $t_{N+1}$ and the projection of $h_N$, inspired by MLP-Speculator~\cite{MLPSpeculatorAccelerating2024wertheimer}:
\begin{gather}
    e_{N+1} = E(t_{N+1}) \\
    s^m = W^m(h_N) + \alpha \cdot e_{N+1}, m \in {f, i, o, c}
\end{gather}
where we define projection $W^f, W^i, W^o, W^c \in \mathbb{R} ^{d_0 \times d}$ for forget, input, output, and cell state. $\alpha$ is defined following \citet{MLPSpeculatorAccelerating2024wertheimer} as follows:
\begin{gather}
\alpha_0 = 2^{-\frac{1}{2n}}\\
\alpha = \frac{2 \alpha_0}{(1-\alpha_0^2) \cdot d}
\end{gather}
where $n$ is the maximum tree depth we aim to generate.

Now we imitate the flow of LSTM forward as follows:
\begin{gather}
g^m = \sigma(s^m), m \in {f, i, o}\\
s^c = f(s^c) \cdot g^i\\
z = z \cdot g^f + s^c\\
h_{N+1} = f(z) \cdot g^o
\end{gather}
where $\sigma$ is the sigmoid function, and $f$ is the layer normalization along with an activation function, which we use GeLU. $z$ is the cell state, which we initialize with zeros.
We speculate $t_{N+2}$ from $h_{N+1}$ using a trainable head, and recurrently speculate the next token in a similar manner, reusing the trainable weight parameters.

\begin{figure*}[]
\centering
\begin{minipage}[t]{.32\textwidth}
    \centering
    \includegraphics[width=\columnwidth]{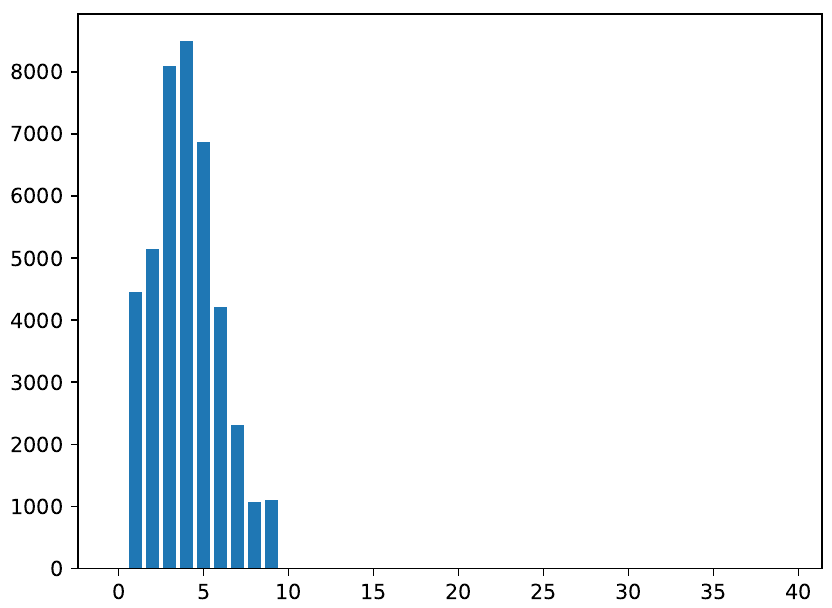}
    \subcaption{Tree decoding (\ours)}\label{fig:ours_tree_hist}
\end{minipage}  
\begin{minipage}[t]{.32\textwidth}
    \includegraphics[width=\textwidth]{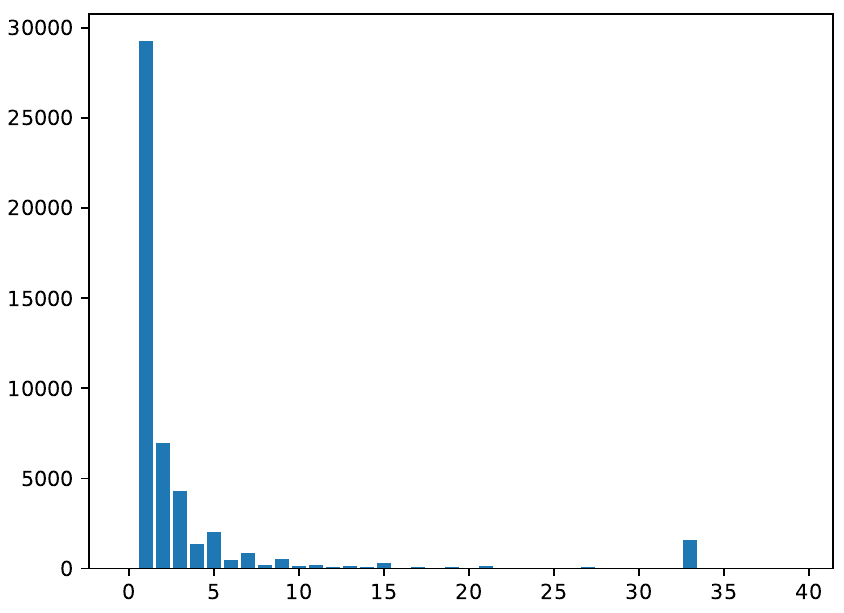}
    \subcaption{Non-tree decoding~\cite{SuffixDecoding2025oliaro}}\label{fig:suffix_non_tree_hist}
\end{minipage}  
\begin{minipage}[t]{.32\textwidth}
    \includegraphics[width=\textwidth]{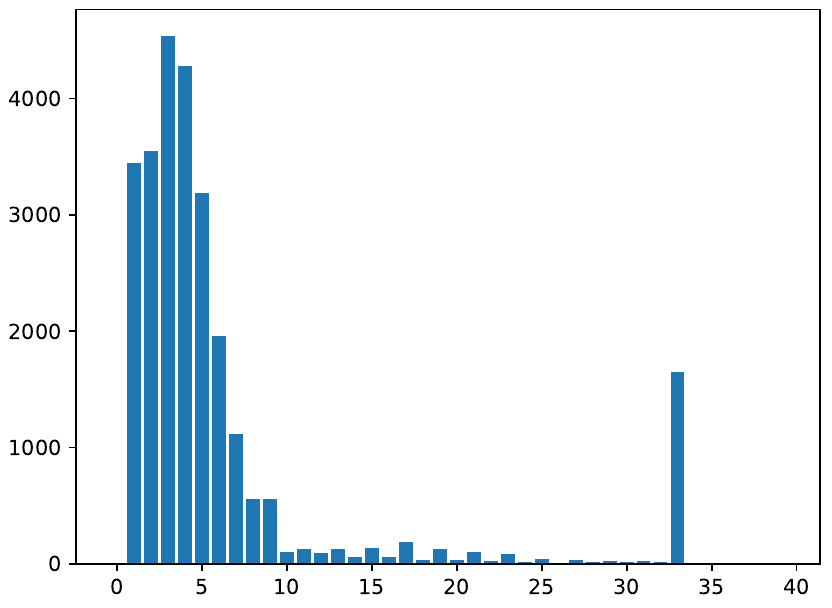}
    \subcaption{Hybrid (\hybridours)}\label{fig:ours_hybrid_hist}
\end{minipage}  
\caption{
Histogram of acceptance length on Llama-3.1-8B-Instruct model on \ourbench. While \ours (left) achieves a higher acceptance length than the non-tree decoding method (middle) on average, the non-tree decoding method often achieves extremely high acceptance length. We build a hybrid method combining the benefits of the two, further improving the average acceptance length (right).}
\label{fig:hist}
\end{figure*}

\subsubsection{\ourtoken in Verifier: Empowering Drafter with Richer Representation}\label{sec:spec_token}
With our simplified architecture, \ours can generalize to long-context inputs. However, the acceptance length is still limited by the drafter's capability.
To empower the drafter with richer representation, we let the target LLM try to estimate an additional token beyond the accepted tokens at the verification step. Then we leverage that estimation in the drafter to increase the acceptance length further.

We describe how to introduce \ourtoken to the verifier in 1) the prefill stage in inference, 2) the decode stage in inference, and 3) the training stage below.

\paragraph{Prefill Stage}
As depicted in \autoref{fig:inference_overview} (a), we append \ourtoken to the input at the prefill stage. This signals the target LLM to generate the hidden state for estimating the additional token. The estimated hidden state is consumed by the drafter to generate draft tokens.

\paragraph{Decode Stage}
At the decode stage, as shown in \autoref{fig:inference_overview} (b), we append \ourtoken after each tree path. We enable this by appending the same number of \ourtoken as the number of tree nodes of the drafted tree, and manipulating the position ids and attention mask accordingly. For example, the first \ourtoken follows the path $(t_{N+1})$, the second \ourtoken follows the path $(t_{N+1},t_{N+2})$, and the third \ourtoken follows the path $(t_{N+1},t_{N+2'})$. When a path is accepted, we take the hidden state of the corresponding \ourtoken as the estimation of the next token. This allows the target LLM to predict beyond the accepted tokens of the drafted tree, providing the drafter richer representation.

Importantly, we keep the computational cost the same as the original tree decoding method by decreasing the tree size by half. In section \ref{sec:exp}, we empirically show that this approach significantly increases the acceptance length even if we pass the same number of tokens to the verifier target LLM.

\paragraph{Training Stage}
To enable signaling with \ourtoken while keeping the target LLM intact, we train the embedding of \ourtoken along with the drafter in an efficient manner.

The efficiency of autoregressive training comes from computing the next token hidden states of all prefixes in a single forward pass. To train the drafter with \ourtoken, we need to compute the hidden states of \ourtoken for all possible prefixes.
As shown in \autoref{fig:training_overview}, we append the same number of \ourtoken as the number of prefixes in the given input. We then manipulate the position ids and attention mask to ensure that each \ourtoken attends only to its preceding tokens. This allows us to compute the hidden states for all possible prefixes in a single forward pass, maintaining training efficiency.

In detail, suppose we get logits $y_{k \oplus 1}, \cdots, y_{k \oplus n}$ speculated from $t_k, h_{k-1}, h_{[SPEC]_{k-1}}$. Then the loss $\mathcal{L} $ is formulated as
\begin{multline}
\mathcal{L} = \frac{1}{N} \sum_k \big( \frac{1}{n} \sum_j CE(y_{k\oplus j}, t_{k+j}) +\\ 
CE(y_{[SPEC]_{k-1}}, t_k)\big)
\end{multline}
where $CE$ is cross-entropy loss, and $y_{[SPEC]_{k-1}}$ is generated from $h_{[SPEC]_{k-1}}$ using the head of the target LLM.

\subsubsection{\hybridours: Hybrid Algorithm with Non-Tree Decoding}\label{sec:hybrid}
While the above architectural innovations for the drafter and verifier significantly improve the acceptance length, we further explore the decoding algorithm itself. We find out the complementary benefit of tree- and non-tree decoding based speculative decoding methods.

As depicted in \autoref{fig:hist}, we find that while some non-tree decoding based methods, such as SuffixDecoding~\cite{SuffixDecoding2025oliaro}, have lower acceptance length than our tree-decoding-based method in average, they can provide extremely high acceptance length in some cases. Since the best case scenario of non-tree decoding is accepting all the sequence of drafted tokens, we conditionally leverage this to enhance our best case scenario.

Algorithm \ref{alg:hybrid} describes \hybridours,
estimating the acceptance length of the non-tree decoding method, the non-tree decoding version of SuffixDecoding~\cite{SuffixDecoding2025oliaro}, as $score$, used for routing decision. We follow \citet{SuffixDecoding2025oliaro} to estimate $score$.

If $score$ is higher than a threshold $c$, we use non-tree decoding method to verify the drafted tokens. To ensure the best case scenario, we do not append any \ourtoken to the input in this case, since doing so would limit the acceptance length when the same computational budget is assumed.

If $score$ is not higher than a threshold $c$, we use our tree-decoding-based method. If the previous step used the non-tree decoding method, we did not use \ourtoken, thus we use \ours trained without \ourtoken.
Otherwise, we use \ours trained with \ourtoken.
In any case of tree-decoding based method, we append \ourtoken to the input and manipulate the position ids and attention mask accordingly, as described in \autoref{fig:inference_overview} (b) before the verification step. Note that this does not alter the non-tree decoding algorithm $SuffixLinear$, since this is only used in the $OurTreeVerify$ step.

In practice, we set $c$ as the largest acceptance length \ours can achieve, to ensure we use \ours in the average case.
As a result, the best-case scenario of non-tree decoding is enabled, while we still leverage the benefits of our tree-decoding-based method in the average-case scenario.

\begin{algorithm}[t]
\small
\caption{Hybrid Decoding Algorithm}\label{alg:hybrid}
\begin{algorithmic}[1]
\Require $l \gets$ Input sequence
\Require $S \gets$ Token id of \ourtoken
\Require $c \gets$ Threshold for switching tree/non-tree decoding
\Require $D_{spec} \gets$ \ours trained with \ourtoken
\Require $D_{nospec} \gets$ \ours trained without \ourtoken
\State $SuffixPrefillCache(l)$
\State $t_{next}, h_{last}, h_{S} = Prefill(l)$ \Comment{Figure 4(a)}
\State $lastwaslinear \gets$ False
\While{End of sequence not reached}
    \State $d, score \gets SuffixLinear(l, t_{next})$
    \If{$score > c$}
        \State $lastwaslinear \gets$ True
        \State $d, t_{next}, h_{last} \gets NonTreeVerify(d)$
    \Else
        \If{$lastwaslinear$}
            \State $d \gets D_{nospec}(t_{next}, h_{last})$
            \State $lastwaslinear \gets$ False
        \Else
            \State $d \gets D_{spec}(t_{next}, h_{last}, h_{S})$
        \EndIf
        \State $Prepare(d)$ \Comment{Append \ourtoken, Attention mask, Position id manipulation}
        \State $d, t_{next}, h_{last}, h_{S} \gets OurTreeVerify(d)$ \Comment{Figure 4(b)}
    \EndIf
    \State $l \gets l;d$ \Comment{sequence concatenation}
    \State $SuffixCache(d)$
    \State $N \gets N + |d|$
\EndWhile
\State \Return $l$
\end{algorithmic}
\end{algorithm}

\begin{table*}[]
\centering
\begin{tabular}{ll|c|c}
\hline
                                &                        & Llama-3.1-8B & Llama-3.3-70B \\ \hline
\multirow{6}{*}{Methods}        & PLD~\cite{PLDPrompt2023saxena}                    & 2.75                  & 2.24                   \\
                                & Suffix Decoding~\cite{SuffixDecoding2025oliaro}                 & 3.41                  & 2.61                   \\
                                & SAMD~\cite{SAM2025hu}                  & 3.18                  & 2.48                   \\
                                & Token Recycling~\cite{Turning2025luo}        & 3.16                  & 2.97                   \\
                                & EAGLE3~\cite{EAGLE32025li}                 & 1.28                  & 1.35                   \\
                                & \ours (ours)       & \textbf{4.00}         & \textbf{4.27}          \\ \hline
\multirow{2}{*}{Hybrid Methods} & SAMD + Token recycling~\cite{SAM2025hu} & 4.98                  & 4.05                   \\
                                & \hybridours (ours) & \textbf{6.14}         & \textbf{5.31}          \\ \hline
\end{tabular}
\caption{Acceptance length comparison on \ourbench with Llama-3.1-8B-Instruct and Llama-3.3-70B-Instruct as the base models.}
\label{tab:accept_length}
\end{table*}
\begin{table}[]
\centering
\setlength{\tabcolsep}{3pt}
\begin{tabular}{ll|c}
\hline
                                &                       & Speedup \\ \hline
\multirow{7}{*}{Methods}        & baseline              & 1.00\texttimes    \\
                                & PLD                   & 1.59\texttimes    \\
                                & Suffix Decoding       & 2.18\texttimes    \\
                                & SAMD                  & 2.16\texttimes    \\
                                & Token Recycling       & 1.75\texttimes    \\
                                & EAGLE3                & 0.81\texttimes    \\
                                & \ours without \ourtoken             & 2.00\texttimes    \\
                                & \ours (ours)             & \textbf{2.35\texttimes}    \\ \hline
\multirow{2}{*}{Hybrids} & SAMD + Token Recyling & 2.77\texttimes    \\
                                & \hybridours (ours)       & \textbf{3.08\texttimes}    \\ \hline
\end{tabular}
\caption{Token generation (tokens/sec) speed comparison on \ourbench with Llama-3.3-70B-Instruct as the base model.}
\label{tab:speedup}
\end{table}

\section{Experiments}\label{sec:exp}
We investigate the following research questions:
\begin{itemize}
    \item \textbf{RQ1:} How does \ours perform on long-context inputs compared to existing speculative decoding methods?
    \item \textbf{RQ2:} Does \ours generally perform well on benchmarks with various context lengths?
    \item \textbf{RQ3:} Does \ourtoken increase acceptance length even if the inference budget is intact?
    \item \textbf{RQ4:} What if we allow a much longer dataset to train a custom EAGLE3-L to increase the context length?
\end{itemize}
\subsection{Experimental Setup}

\paragraph{Target LLMs}
We evaluate on two different scales of LLMs, Llama-3.1-8B-Instruct~\citep{Llama3Llama2024dubey} and Llama-3.3-70B-Instruct~\citep{Llama3Llama2024dubey}, which are used in related works~\cite{EAGLE32025li,SuffixDecoding2025oliaro}.

\paragraph{Comparisons}
We compare our method with various existing speculative decoding methods, including PLD~\citep{PLDPrompt2023saxena}, SuffixDecoding~\citep{SuffixDecoding2025oliaro}, SAMD~\citep{SAM2025hu}, Token Recycling~\citep{Turning2025luo}, and EAGLE3~\citep{EAGLE32025li}. We also compare with hybrid methods, combining SAMD and Token Recycling~\citep{SAM2025hu}. We detail the inference setups in the Appendix.

\paragraph{Training Details}
We train \ours on Ultrachat-200k\footnote{hf.co/datasets/HuggingFaceH4/ultrachat\_200k}~\cite{UltrachatEnhancing2023ding} and Magicoder\footnote{hf.co/datasets/ise-uiuc/Magicoder-OSS-Instruct-75K}~\cite{Magicoder2024wei}. Inspired by \citet{MLPSpeculatorAccelerating2024wertheimer}, we first chunk the data by size of 64 and generate 256 tokens in the preprocessing step, using vLLM~\cite{vllmEfficient2023kwon}. We then train \ours with these generated chunks with sequence length of 256. We train with batch size of 2048, learning rate of 1e-3, for 3000 iterations. We use hidden size $d$ of 12288. All training is done with 8\texttimes H200 GPUs.

To train the long-context version of EAGLE3, we leverage SpecForge~\cite{specforge2025}, following their default settings, except for setting the number of epochs as 16 to give a fair training time as \ours. We set the maximum sequence length as 32K, since we couldn't increase more due to the OOM error. 

\subsection{RQ1: \ours is the Best on Long-Context Inputs over Various Model Sizes}
Among various comparisons, \ours achieves the highest acceptance length on \ourbench, for both Llama-3.1-8B-Instruct and Llama-3.3-70B-Instruct (\autoref{tab:accept_length}). Surprisingly, EAGLE3 shows an acceptance length of around 1.28, which is far below the training-free methods, such as PLD, Suffix Decoding, SAMD, or Token Recycling.
In contrast, \ours achieves an acceptance length of around 4.00-4.27, which is way beyond existing methods.

Moreover, \ours shows better acceptance length on larger models, such as Llama-3.3-70B-Instruct. This is different from the trend of training-free retrieval methods such as PLD, Suffix Decoding, SAMD, and Token Recycling, whose acceptance length gets lower. We hypothesize that as the output distribution becomes more delicate, training-free methods suffer from predicting the complicated output distribution.

When we compare the hybrid methods, \hybridours achieves an acceptance length of 6.14 at most, almost 5\texttimes \xspace than that of EAGLE3. Since we combine \ours with the training-free retrieval method, Suffix Decoding, the degradation in the larger model is expected.

These improvement translates into the highest token generation speed as well (\autoref{tab:speedup}). Surprisingly, the state-of-the-art method, EAGLE3, makes the generation speed even slower, to 0.81\texttimes \xspace, due to their small acceptance length and drafting overhead. In contrast, \ours and \hybridours neatly show the best speedup, compared with other speculative decoding baselines.
 
\subsection{RQ2: \ours Generlizes over Various Context Length}
\autoref{tab:specbench} shows similar acceptance length over benchmarks with different length distribution, while others do not. For example, while EAGLE3 shows the best acceptance length on the short benchmark, SpecBench, it shows the worst acceptance length on the long benchmark, \ourbench. Retrieval methods, such as PLD, Suffix Decoding, SAMD, and Token Recycling show much lower acceptance length on the short benchmark, SpecBench, compared with the longer benchmark, \ourbench.
In contrast, the acceptance length of \ours is above 4 on both benchmarks.

\begin{table}[]
\centering
\setlength{\tabcolsep}{1pt}
\begin{tabular}{l|cc|c}
\hline
                & SpecBench & \ourbench & min \\ \hline
PLD             & 1.41      & 2.75      & 1.41\\
Suffix Decoding & 1.56      & 3.41      & 1.56\\
SAMD            & 1.42      & 3.18      & 1.42\\
Token Recycling & 2.73      & 3.16      & 2.73\\
EAGLE3          & 5.79      & 1.28      & 1.28\\ 
\ours           & 4.14      & 4.00      & \textbf{4.00}\\\hline
\end{tabular}
\caption{Acceptance length comparison on SpecBench~\citep{SpecBenchUnlocking2024xia} and \ourbench with Llama-3.1-8B-Instruct as the base model. The acceptance length of \ours is consistent across benchmarks with various length.}
\label{tab:specbench}
\end{table}

\subsection{RQ3: \ourtoken Significantly Improves Acceptance Length}
\begin{figure}[]
\centering
\includegraphics[width=0.8\linewidth]{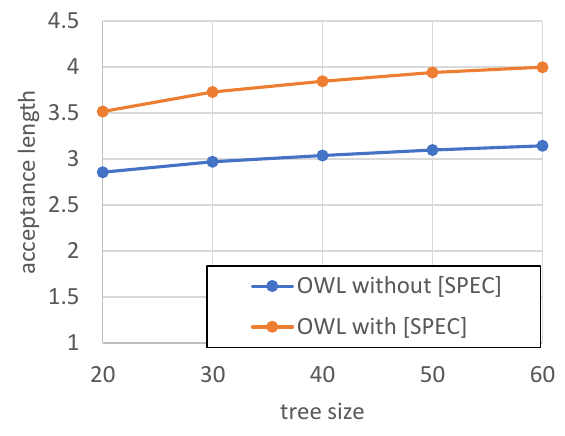}
\caption{Acceptance length comparison between \ours with and without \ourtoken, with same tree size is set.}
\label{fig:ourtoken_length}
\end{figure}

When we compare passing the same number of tokens to the verifier, with and without \ourtoken, including \ourtoken in the tokens significantly increases acceptance length (\autoref{fig:ourtoken_length}). For example, if we generate 30 tokens from the drafter and append 30 \ourtoken s to make a tree size of 60, this increases acceptance length by almost 1, compared with generating 60 tokens from the drafter and directly passing to the verifier. The acceptance length is increasing more rapidly as we increase the tree size, benefiting from tree-decoding more.

Moreover, the row \ours without \ourtoken and \ours in \autoref{tab:speedup} shows that \ourtoken contributes to the actual speedup as well.
These verify that \ourtoken significantly improves speedup without affecting the latency.

\subsection{RQ4: Even if We Train Longer-Context EAGLE3-L, \ours Outperforms as Well}
\begin{table}[]
\centering
\setlength{\tabcolsep}{1.5pt}
\begin{tabular}{l|ccc}
\hline
                     & \ours          & EAGLE3 & EAGLE3-L \\ \hline
train context window & \textbf{256}   & 2048   & 32768               \\
train data length       & \textbf{short} & short  & long                \\
acceptance length    & \textbf{4.00}  & 1.28   & 3.23                \\ \hline
\end{tabular}
\caption{\ourbench comparison of \ours, EAGLE3, and EAGLE3-L which is a version we trained with a much longer context window to construct a stronger baseline.}
\label{tab:longeagle}
\end{table}
Even if we train an EAGLE3-L by allowing much longer data explicitly to support a longer context window, \ours outperform it (\autoref{tab:longeagle}). 
We use LongAlign~\cite{LongAlign2024bai} and LongWriter~\cite{LongWriter2025bai} to support such a long sequence to train the speculator.
\ours is more efficient-- it does not require a special dataset, or longer sequence length, while achieving higher acceptance length.

\subsection{Ablation Studies}
\begin{table}[]
\centering
\setlength{\tabcolsep}{1pt}
\begin{tabular}{l|ccc|c}
\hline
                        & LSTM      & \ourtoken & Hybrid    & AL   \\ \hline
\hybridours             & \ding{51} & \ding{51} & \ding{51} & 6.14 \\
\ours                   & \ding{51} & \ding{51} &           & 4.00 \\
\ours w/o \ourtoken & \ding{51} &           &           & 3.14 \\
RNN-based               &           &           &           & 2.99 \\
EAGLE3                  &           &           &           & 1.28 \\ \hline
\end{tabular}
\caption{Contribution of each innovation to the acceptance length (AL) with Llama-3.1-8B-Instruct on \ourbench. RNN-based follows the architecture of \citet{MLPSpeculatorAccelerating2024wertheimer}.}
\label{tab:ablation}
\end{table}
\autoref{tab:ablation} shows that each innovation contributes to the performance of \ours. While we arbitrarily used an LSTM architecture to remove window length-dependency, we also implement an RNN-based architecture, following \citet{MLPSpeculatorAccelerating2024wertheimer}. LSTM architecture is better, \ourtoken makes it even better, and \hybridours achieves the best.

\section{Conclusion}
In this paper, we identify the limitations of existing speculative decoding methods on long-context inputs, and propose \ours, a length-generalized speculative decoding method for long-context inputs. Our method can achieve almost 5\texttimes \xspace higher acceptance length than EAGLE3 on long-context inputs. 
We publicly release our code and datasets to facilitate future research.

\section{Limitations}
While \ours shows the best performance on long-context inputs, real-world workload may contain only a small batch of short-context samples as well. For this less-likely scenario, we can consider a hybrid with EAGLE3, which we leave as future work.

Although we emperically showed that LSTM architecture works well for legnth-generalization, other alternatives, such as state-space models~\cite{S4Efficiently2022gu}, are yet unexplored. We leave those directions as future work.

\bibliography{2026ICLR_autogen.bib}

\appendix
\begin{figure*}[]
\centering
\begin{minipage}[t]{.35\textwidth}
    \includegraphics[width=\textwidth]{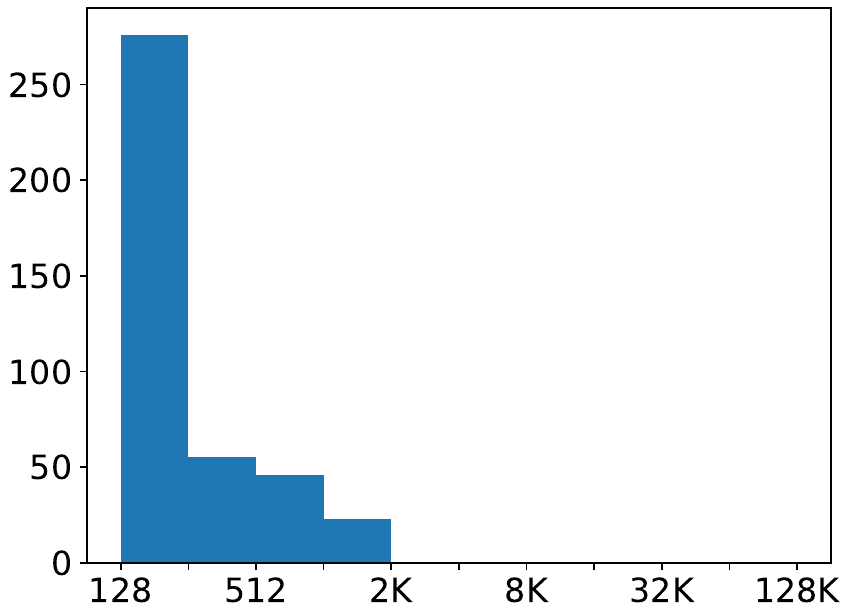}
    \subcaption{EAGLE3 benchmark~\citep{EAGLE32025li}}\label{fig:eagle3_len}
\end{minipage}  
\begin{minipage}[t]{.35\textwidth}
    \includegraphics[width=\textwidth]{figures/wildchat_eq_multiturn_64K_Meta-Llama-3.1-8B-Instruct_lens.pdf}
    \subcaption{\ourbench (ours)}\label{fig:ourbench_len2}
\end{minipage}  
\caption{
Context length distribution of benchmark used in EAGLE3~\citep{EAGLE32025li} and \ourbench.}
\label{fig:len_eagle3_ourbench}
\end{figure*}
\section{Appendix}
\subsection{Inference Details}
We follow implementations in SpecBench~\cite{SpecBenchUnlocking2024xia}, and follow their recommended hyperparameters, except for increasing the tree depth of EAGLE3 to 8, following their paper~\cite{EAGLE32025li}. 
In detail, we used hyperparameters as follows:
\begin{itemize}
    \item EAGLE3: We use tree size of 60, top-k of 10, and depth of 8. We use official models provided by the author for Llama-3.1-8B-Instruct\footnote{hf.co/yuhuili/EAGLE3-LLaMA3.1-Instruct-8B} and Llama-3.3-70B-Instruct.\footnote{hf.co/yuhuili/EAGLE3-LLaMA3.3-Instruct-70B}
    \item \ours: Following EAGLE3, we use tree size of 60, top-k of 10, and depth of 8.
    \item PLD: We use max-ngram-size of 3, and num-pred-tokens of 10.
    \item SAMD: We use n-predicts of 40, max-predicts of 70. For hybrid, we use len-threshold of 5, len-bias of 5.
    \item Token recycling: We use output-id-topk of 8, with tree version 2.2.2.
    \item Suffix decoding: We use max-spec-factor of 2. For hybrid, we use max-suffix-depth of 64, and suffix-threshold of 9, the maximum acceptance length \ours can achieve (\S \ref{sec:hybrid}).
\end{itemize}

Following related works~\cite{EAGLE32025li,SuffixDecoding2025oliaro,SpecBenchUnlocking2024xia,Turning2025luo,SAM2025hu}, we use batch size of 1. We fairly optimized each methods with a static cache design provided by SAMD~\cite{SAM2025hu}, and use fp16 data type.
For Llama-3.1-8B-Instruct, we use 1\texttimes H200 GPU, and for Llama-3.3-70B-Instruct, we use 8\texttimes H200 GPUs.

\subsection{Length Distribution of EAGLE3 Benchmark and \ourbench}
\autoref{fig:len_eagle3_ourbench} further compares the length distribution between the benchmark EAGLE3 used and \ourbench.

\end{document}